# Towards *Physarum* robots: computing and manipulating on water surface

Andrew Adamatzky

Computing, Engineering and Mathematical Sciences, University of the West of England, Bristol, United Kingdom

and

Bristol Robotics Laboratory, Bristol, United Kingdom

andrew.adamatzky@uwe.ac.uk

**Abstract**

Plasmodium of *Physarym polycephalum* is an ideal biological substrate for implementing concurrent and parallel computation, including combinatorial geometry and optimization on graphs. We report results of scoping experiments on *Physarum* computing in conditions of minimal friction, on the water surface. We show that plasmodium of *Physarum* is capable for computing a basic spanning trees and manipulating of light-weight objects. We speculate that our results pave the pathways towards design and implementation of amorphous biological robots.

## Introduction

Plasmodium, the vegetative stage of slime mould *Physarum polycephalum*, is a single cell, with thousands of diploid nuclei, formed when individual flagellated cells or amoebas of *Physarum polycephalum* swarm together and fuse. The plasmodium is visible by naked eye. When plasmodium is placed on an appropriate substrate, the plasmodium propagates and searches for sources of nutrients (bacteria). When such sources are located and taken over, the plasmodium forms veins of protoplasm. The veins can branch, and eventually the plasmodium spans the sources of nutrients with a dynamic proximity graph, resembling, but not perfectly matching graphs from the family of *k*-skeletons [Kirkpatrick and Radke, 1985].

A large size of the plasmodium allows the single cell to be highly amorphous. The plasmodium shows synchronous oscillation of cytoplasm throughout its cell body, and oscillatory patterns control the behaviours of the cell. All the parts of the cell behave cooperatively in exploring the space, searching for nutrients and optimizing network of streaming protoplasm. Due to its unique features and relative ease of experimentation with, the plasmodium became a test biological substrate for implementation of various computational tasks. The problems solved by the plasmodium include maze-solving [Nakagaki, 2000, 2001], calculation of efficient networks [Nakagaki 2003, 2004; Shirakawa & Gunji, 2007], construction of logical gates [Tsuda et al, 2004], formation of Voronoi diagram [Shirakawa & Gunji, 2007a], and robot control [Tsuda et al., 2007].

The oscillatory cytoplasm of the plasmodium can be seen as a spatially extended nonlinear excitable media [Matsumoto et al., 1988; Nakagaki et al., 1999; Yamada et al., 2007]. In our previous papers we hypothesized that the plasmodium of *Physarum* is a biological analogue of a chemical reaction-diffusion system encapsulated in an elastic and growing membrane [Adamatzky, 2007a]. Such an encapsulation enables the plasmodium to function as a massively-parallel reaction-diffusion computer [Adamatzky, De Lacy Costello, Asai, 2005] and also to solve few tasks which reaction-diffusion computers could not do, e.g. construction of spanning trees [Adamatzky, 2007], and implementation of storage modification machines [Adamatzky, 2008]. Also, under certain experimental conditions, the plasmodium exhibits travelling self-localizations, implements collision-based logical circuits and thus is capable for universal computation [Adamatzky, De Lacy Costello, Shirakawa, 2008].

Being encapsulated in an elastic membrane the plasmodium can be capable of not only computing over spatially distributed data-sets but also physically manipulating elements of the data-sets. If a sensible, controllable and, ideally, programmable movement of the plasmodium and manipulation by the plasmodium could be achieved, this would open ways for experimental implementations of amorphous robotic devices. There are already seeds of an emerging theory of artificial amoeboid robots [Yokoi and Kakazu, 1992; Yokoi et al., 2003; Shimizu and Ishiguro, 2008].

In present paper we undertook a set of scoping experiments on establishing links between *Physarum* computing and *Physarum* robotics. We have chosen water surface as a physical substrate for the plasmodium development to study how topology of the plasmodium network can be dynamically updated, without being stuck to a non-liquid substrate, and how small objects floating on the surface can be manipulated by the plasmodium's pseudopodia.

**Methods**

Plasmodium of *Physarum polycephalum* was cultivated on a wet paper towels in dark ventilated containers, oat flakes were supplied as a substrate for bacteria on which the plasmodium feeds. We used several test arenas for observing behaviour of the plasmodium and scoping experiments on plasmodium-induced manipulation of floating objects. These are Petri dishes with base diameters 20 mm and 90 mm, and rectangular plastic containers 200 by 150 mm. The dishes and containers were filled by 1/3 with distilled water. Data points, to be spanned by the plasmodium, were represented either by 5-10 mm sized pieces of a plastic foam, which were either fixed to bottom of Petri dishes or left floating on water surface (in case of large containers). Oat flakes were placed on top of the foam pieces. Foam pieces, where plasmodium was initially placed, and the pieces with oat flakes were anchored to bottom of containers. Tiny foam pieces to be manipulated by plasmodium were left free-floating.

**Results: Computing and Manipulating**

To demonstrate that a substrate is suitable for robotics implementations one must demonstrate that the substrate is capable for sensing of environment and responsiveness to external stimulus, solving complex computational tasks on spatially distributed data sets, locomotion, and manipulation of objects. We provide basic demonstrations which may indicate that plasmodium of *Physarum polycephalum* can be successfully used in future experiments on laboratory implementations of amorphous biological robots.

Surface of water is in tension therefore it physically supports propagating plasmodium, when its contact weight to contact area ratio is small. When placed in an experimental container the plasmodium forms pseudopodia aimed to search for sources of nutrients. In most experiments 'growth part' of the pseudopodia has tree-like structure for fine detection of chemo-gradients in the medium, which also minimized weight to area ratio. Examples of tree-like propagating pseudopodia are shown in Fig. 1.

In Fig. 1b we can see that not always pseudopodia grow towards source of nutrients, there is a pseudopodia growing South-West, where no sources of nutrients located. This happens possibly because in large-sized containers volume of air is too large to support a reliable and stationary gradient of chemo-attractants. This may pose a difficulty for the plasmodium to locate and span all sources of nutrients in large-sized containers.

In Petri dishes volume of air is small and, supposedly the air is stationary, therefore plasmodium easily locates sources of nutrients (Fig. 2). It thus builds spanning trees where graph nodes (to be spanned) are presented by pieces of foam with oat flakes on top. In Fig. 2 we can see that originally the plasmodium was positioned at the Southern domain. In twelve hours the plasmodium builds a link with Western domain, and then starts to propagate pseudopodia to the Eastern domain.

When the plasmodium spans sources of nutrients, it produces many 'redundant' branches (Fig. 2). These branches of pseudopodia are necessary for space exploration but do not represent minimal edges connecting the nodes of the spanning tree. These 'redundant' branches are removed at later stages of the spanning tree development. See a well-established spanning tree of data-points in Fig. 3. Initially the plasmodium was placed in the Western domain, and the plasmodium has constructed the spanning tree in 15 hours.

We demonstrated that the plasmodium does explore space and computes a spanning tree on the water surface, when placed initially on one of the floating objects. Would the plasmodium be as well operational when placed just on the surface of the water? As we can see in Fig. 4 the plasmodium works perfectly. We placed a piece of plasmodium on bare surface of water (Fig. 4, start). In three hours the plasmodium forms an almost circular front of propagating pseudopodia, which reach two stationary domains with oat reaches in eight hours (Fig. 4).

In usual conditions (on a wet solid or gel substrate) edges of spanning trees, presented by protoplasmic tubes, adhere to the surface of the substrate [Nakagaki et al., 2003; Adamatzky, 2007, 2008]. Therefore the edges cannot move, and the only way the plasmodium can do a dynamical update is to make a protoplasmic tube inoperative and to form a new edge instead (membrane shell of the ceased link will remain on the substrate, e.g. see [Adamatzky, 2008]). When plasmodium operates on water surface, cohesion between the water surface and membrane of protoplasmic tubes is small enough for the protoplasmic tubes to move freely. Thus the plasmodium can make the tubes almost straight and thus minimize costs of the transfer and communication between its distant parts. Two examples of the straightening of the protoplasmic tubes are shown in Fig. 5. Such straightening is a result of the tubes' becoming shorter due to contraction.

Presence of a contraction may indicate that if two floating objects (both with sources of nutrients) are connected by a protoplasmic tube then the objects will be pulled together due to shortening of the protoplasmic tube. We did not manage to demonstrate this exact phenomenon of pulling two floating objects together, however we got experimental evidence of pushing and pulling of single floating objects by the plasmodium's pseudopodia. The plasmodium-induced pushing and pulling are exemplified in Figs. 6 and 7.

To demonstrate pushing we placed a very light-weight piece of plastic foam on the water surface nearby the plasmodium (Fig. 6, 0 hours). The plasmodium develops a pseudopodium which propagates towards the light-weight piece of foam (Fig. 6, 5 hours). Due to gravity force acting on the pseudopodia a ripple is formed on the water surface (Fig. 6, 9 hours), which causes pushes the piece of foam away from the growing pseudopodia's tip (Fig. 6, 13 hours). Due to absence of any nutrients on the pushed piece of foam, the plasmodium abandons its attempt to occupy the piece and retracts the pseudopodia (Fig. 6, 16 hours). The piece remains stationary: it became shifted away from its original position.

In the second example, Fig. 7, we observe pulling of the light-weight object. The piece of foam to be pulled is placed between two anchored objects (Fig. 7, 0 hours). One object hosts the plasmodium another object has an oat flake on top (i.e. attracts the plasmodium). A pseudopodium grows from the plasmodium's original location towards the site with the source of nutrients. The pseudopodium occupies the piece of foam (Fig. 7, 15 hours) and then continues its propagation towards the source of nutrients. When the source of nutrients is reached (Fig. 7, 22 hours) the protoplasmic tubes connecting two anchored objects contract and straighten thus causing the light-weight object to be pulled towards the source of nutrients (Fig. 7, 32 hours). The pushing and pulling capabilities of the plasmodium can be utilized in constructions of water-surface based distributed manipulators [Hosokawa et al., 1996; Adamatzky et al., 2005].

**Discussion**

Inspired by biomechanics of surface walking insects, see e.g. [McAlister, 1959; Suter et al., 1997, Suter, 1999; Suter and Wildman, 1999], our previous studies on implementation of computing tasks in the plasmodium [Adamatzky, 2007, 2008], and our ideas on design and fabrication of biological amorphous robots [Kennedy et al., 2001] we decided to explore operational capabilities of plasmodium of *Physarum polycephalum* on the water surface. We were interested to demonstrate that the plasmodium possesses the essential features of distributed robotics devices: sensing, computing, locomotion and manipulation.

Why water surface? We have chosen water surface as an experimental substrate because there is minimal friction and cohesions between the plasmodium's pseudopodia and the surface, always near ideal humidity for the plasmodium, continuous removal of metabolites and excretions from the plasmodium's body and protoplasmic tubes, increases chances of achieving manipulation of objects by the plasmodium.

We demonstrated experimentally that the plasmodium (1) senses data-objects represented by sources of nutrients, (2) calculates shortest path between the data-objects, and approximates spanning trees where the data-objects are nodes (in principle, a spanning tree of slowly moving data-objects can be calculated as well), (3) pushes and pulls light-weight objects placed on the water surface. The findings indicated that the plasmodium of *Physarum polecephalum* is a perspective candidate for the role of spatially extended robots implemented on biological substrates.

Our experiment also show that Physarum implementation of Kolmogorov-Uspensky machine [Adamatzky, 2008] can be extended to mechanical version of the storage modification machine by adding PUSH NODE and PULL NODE operations. To translocate nodes selectively in the storage structures we may need to assign certain attributes. This can be done by marking nodes with different species of colors; in [Adamatzky, 2008] we demonstrated that the plasmodium exhibits strong preferences to certain food colouring, is neutral to others, and that some food colourings repel the plasmodium. Such preference hierarchy can be mapped onto the mobile data storage structure.

More future experiments is required indeed to develop ideas derived from our scoping experiments to the full working prototypes of the *Physarum* robots and mechanical Kolmogorov-Uspenski machines.

# FIGURES

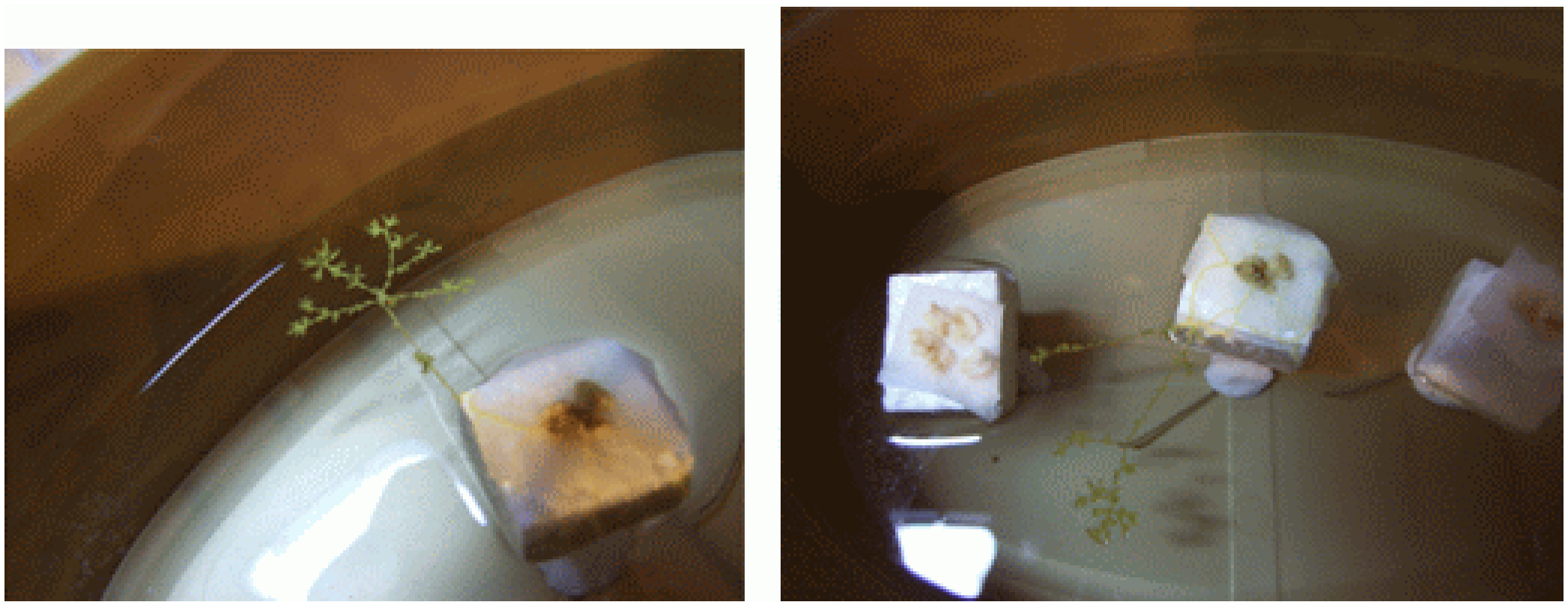

a b

Fig. 1. The plasmodium explores experimental arena by propagating tree-like pseudopodia.

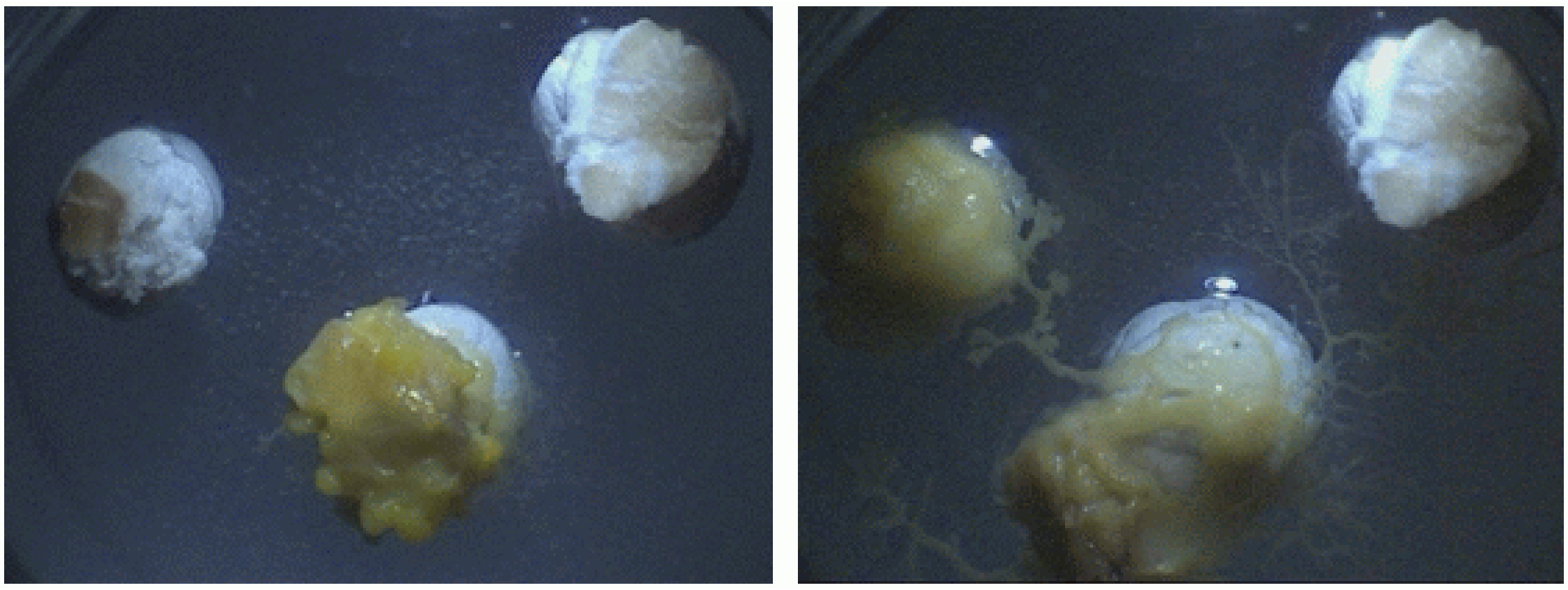

Start 12 hours

Fig. 2. Plasmodium builds links connecting its original domain of residence with two new sites.

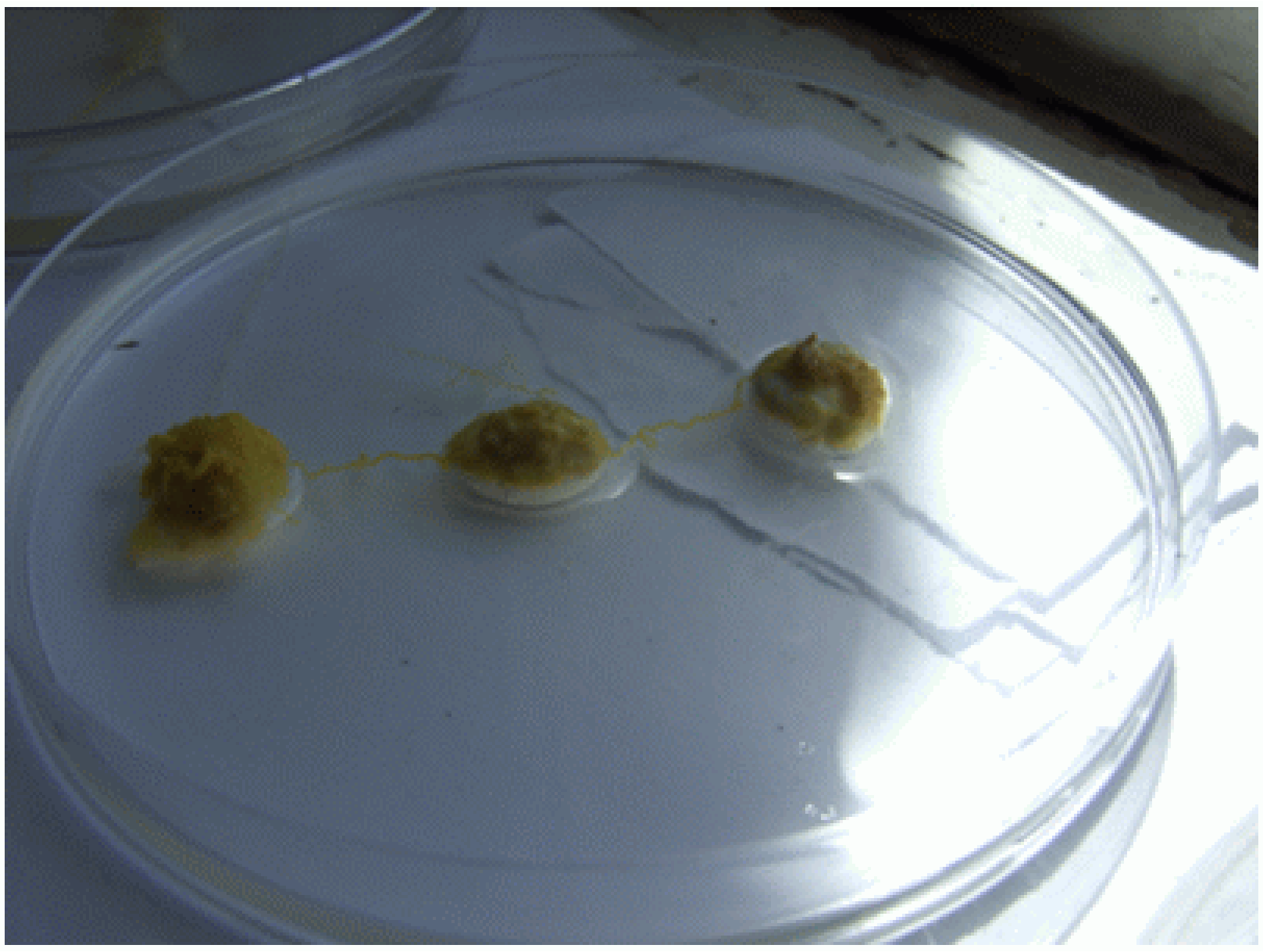

Fig.3. Spanning tree of three points constructed by the plasmodium.

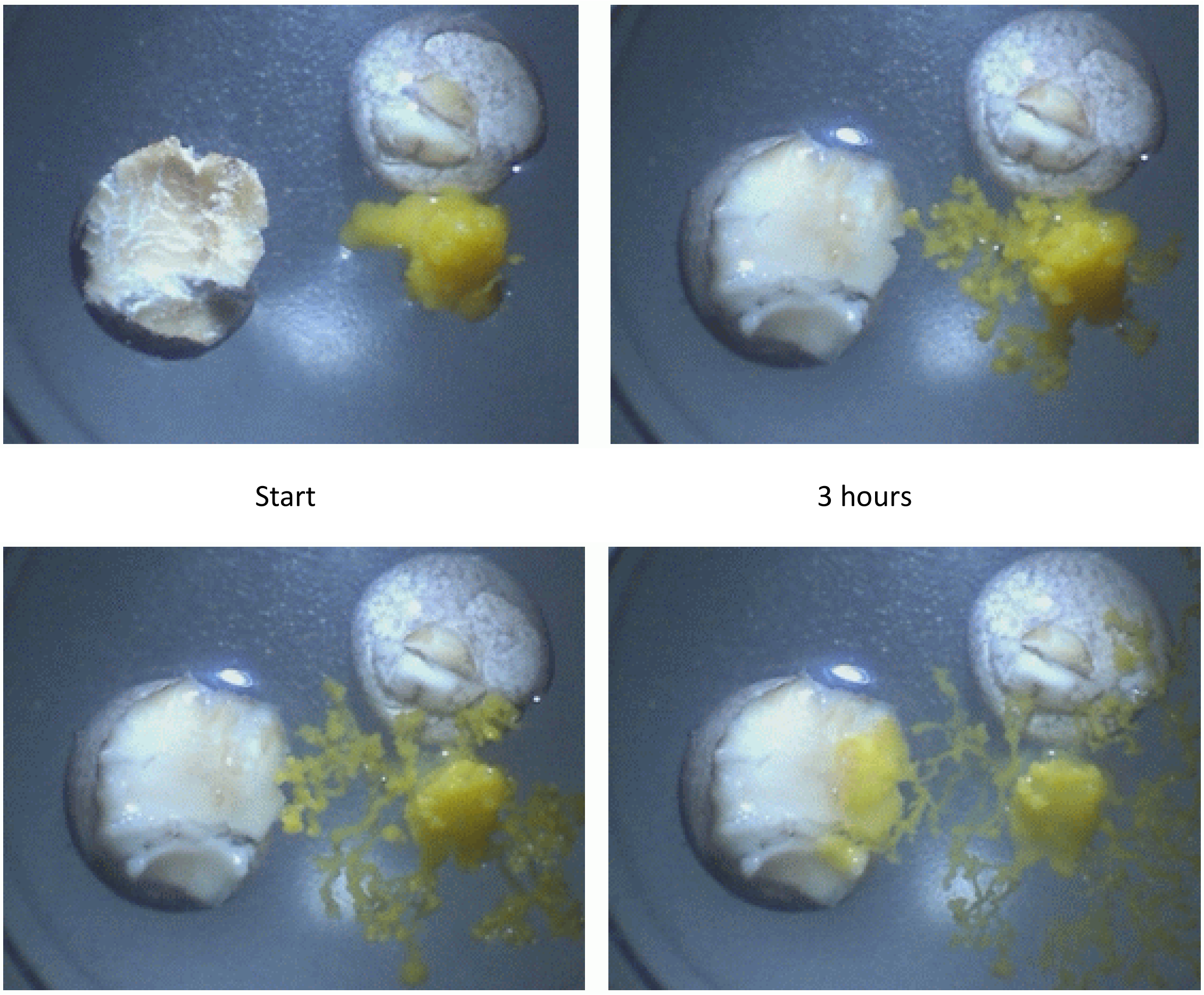

Start 3 hours

5 hours 8 hours

Fig. 4. Plasmodium starts its development on the water surface and occupies two sources of nutrients.

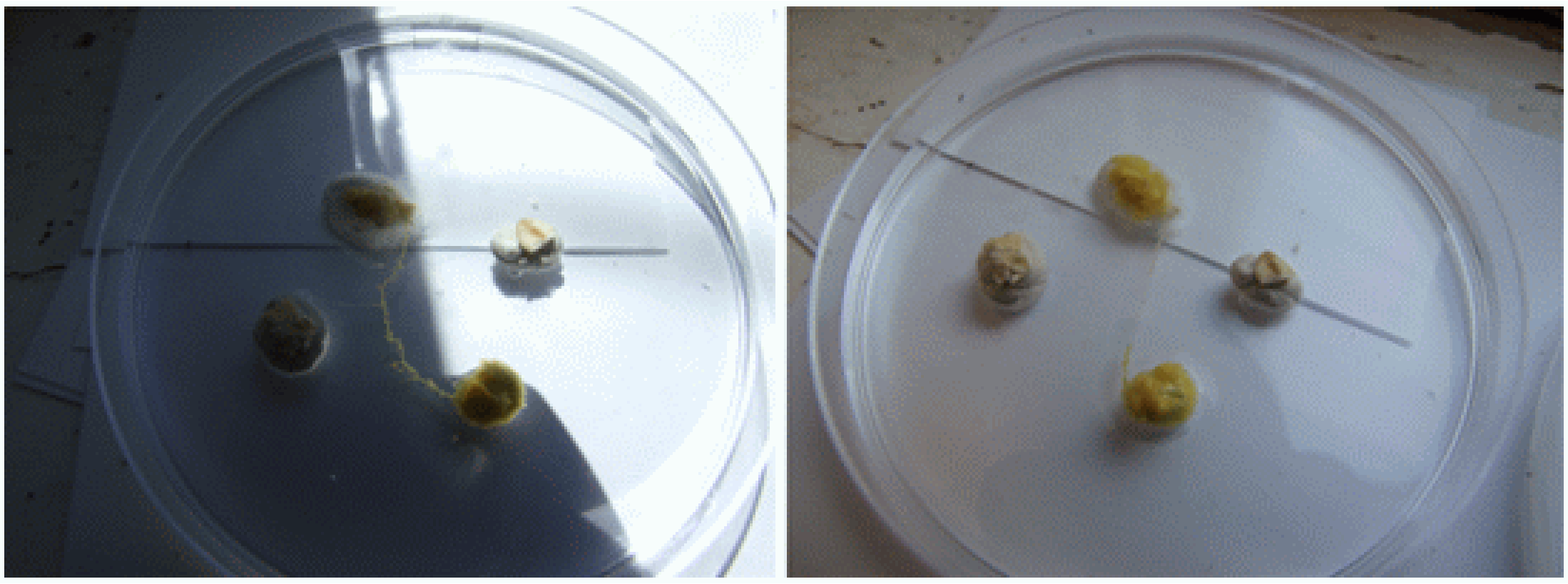

a b (12 hours later)

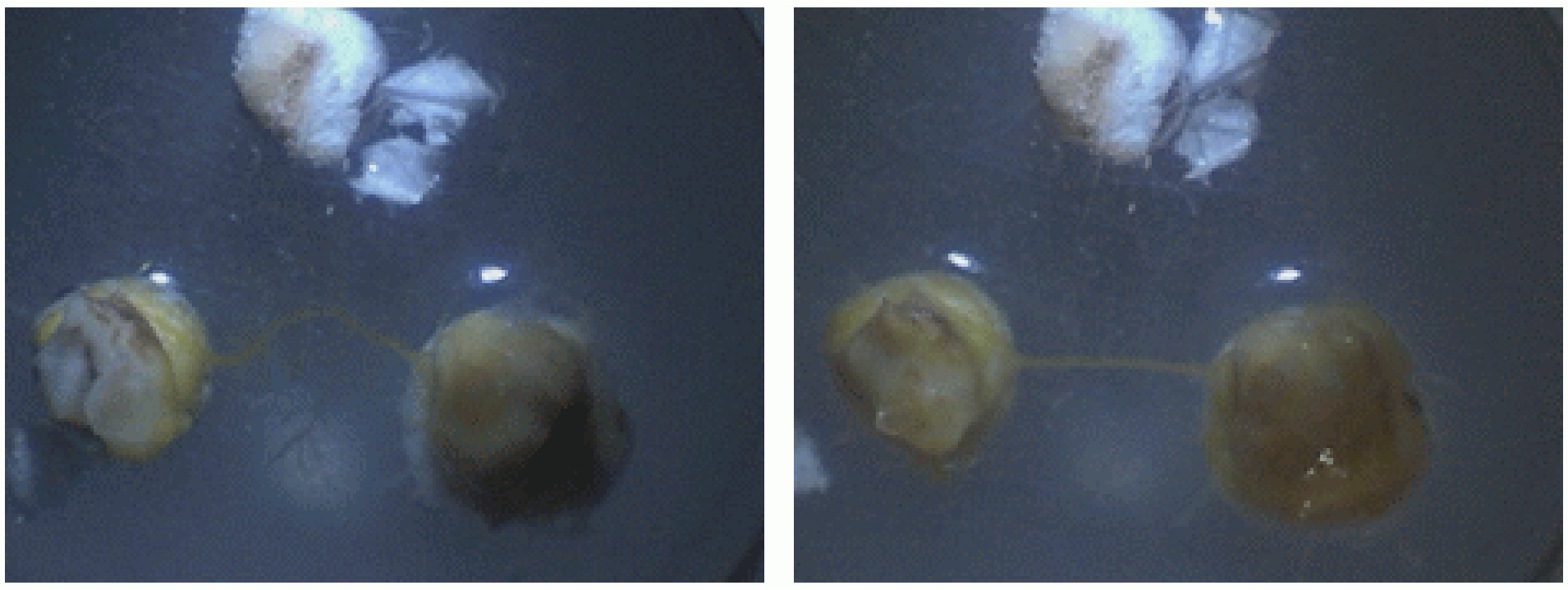

c d (12 hours later)

Fig. 5. Examples of straightening of protoplasmic tubes. In photographs (a) and (c) tubes are longer than necessary. In photographs (b) and (d) the tubes correspond to minimal shortest path between the sites they are connecting.

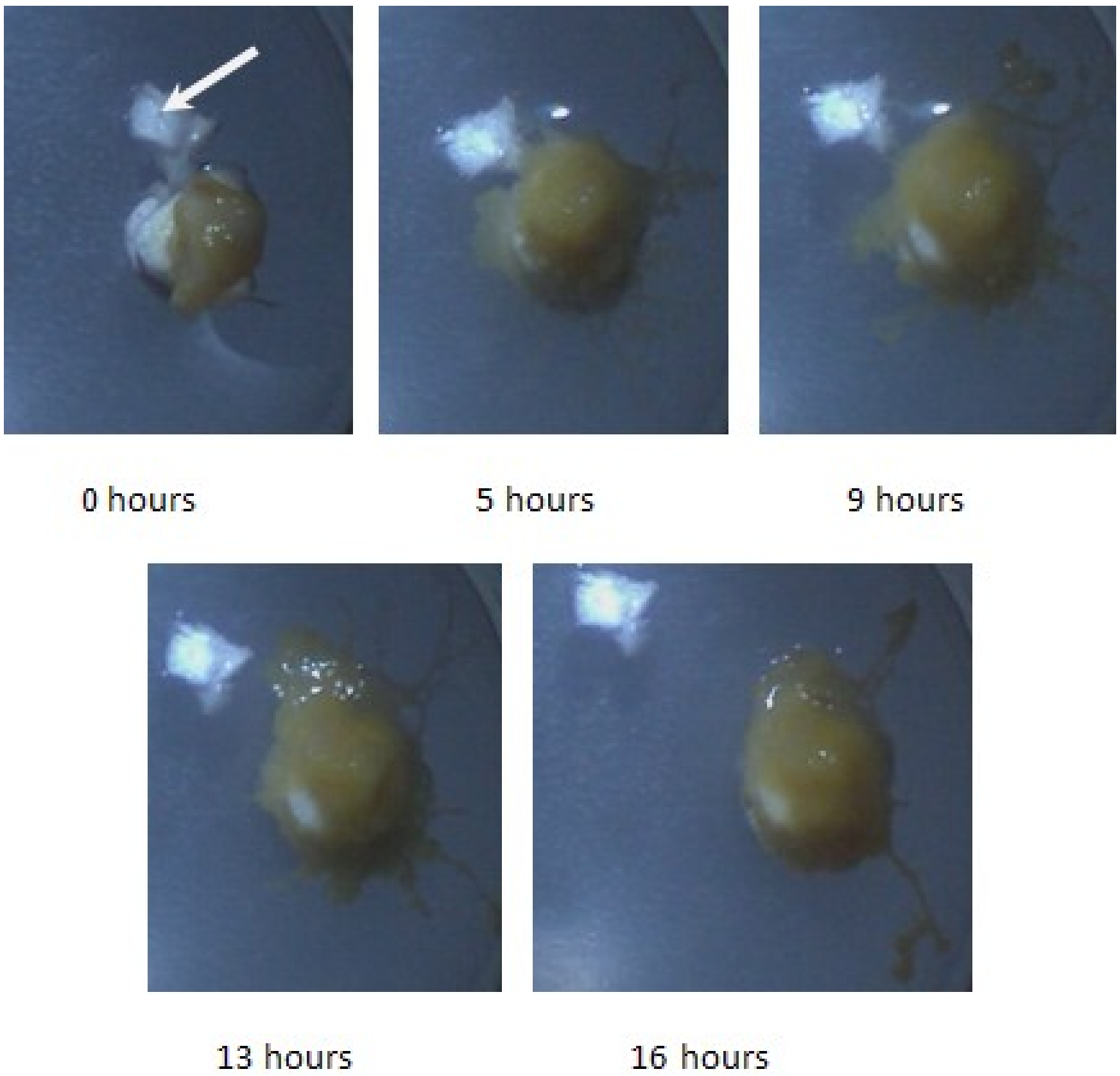

Fig. 6. Photographs demonstrate that the plasmodium can push light-weight floating objects. The object to be pushed is indicated by white arrow in the first photograph the series.

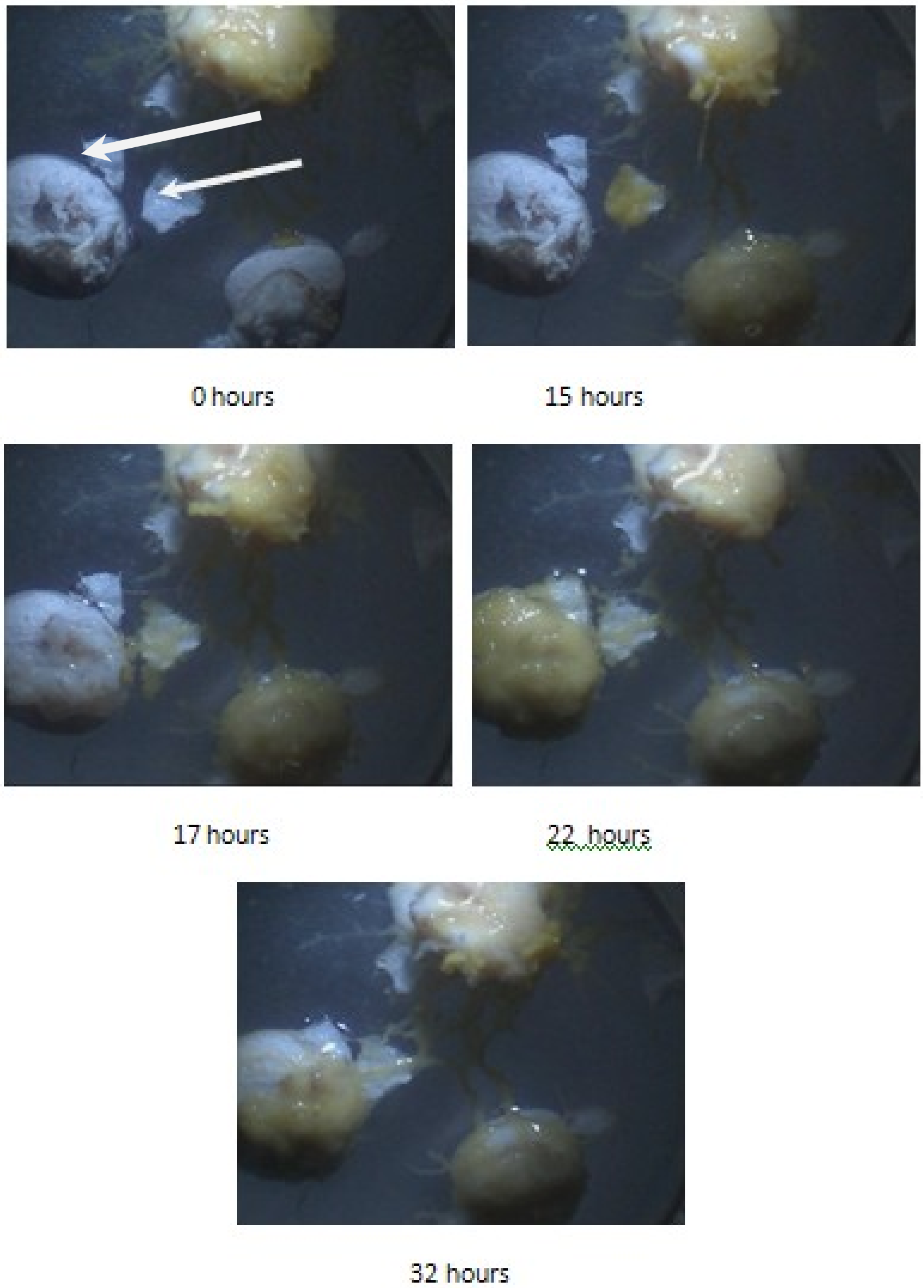

Fig. 7. Photographs demonstrate that the plasmodium can pull lightweight object. The object to be pulled is indicated by white arrow in the first photograph.